\documentclass[conference]{IEEEtran}
\IEEEoverridecommandlockouts
\usepackage[left=0.625in,right=0.625in,top=0.75in,bottom=1in,columnsep=0.202in]{geometry}
\usepackage{cite}
\usepackage{amsmath,amssymb,amsfonts}
\usepackage{algorithmic}
\usepackage{graphicx}
\usepackage{textcomp}
\usepackage{xcolor}
\usepackage{algorithmic}
\usepackage{graphicx}
\usepackage{textcomp}
\usepackage{xcolor}
\usepackage{siunitx}
\usepackage{subfigure}
\usepackage{booktabs}

\usepackage{amsmath,amsthm,amssymb,amsfonts}
\usepackage{graphicx}
\usepackage{epstopdf}

\usepackage{array,booktabs,multirow,makecell,tabularx,tikz}
\usepackage{colortbl}

\usepackage{caption,subcaption}
\usepackage{fancyhdr}
\usepackage{float}
\usepackage{stmaryrd}
\usepackage{dsfont}
\usepackage{color}

\makeatletter
\def\ps@plain{%
  \let\@mkboth\@gobbletwo%
  \let\@oddhead\@empty%
  \let\@evenhead\@empty%
  \let\@oddfoot\@empty%
  \let\@evenfoot\@empty%
}

\def\ps@headings{%
  \let\@mkboth\markboth%
  \let\@oddhead\@empty%
  \let\@evenhead\@empty%
  \let\@oddfoot\@empty%
  \let\@evenfoot\@empty%
}

\def\ps@IEEEtitlepagestyle{%
  \let\@mkboth\@gobbletwo%
  \let\@oddhead\@empty%
  \let\@evenhead\@empty%
  \let\@oddfoot\@empty%
  \let\@evenfoot\@empty%
}
\makeatother

\allowdisplaybreaks[4]

\definecolor{headercolor}{RGB}{52, 101, 164}
\definecolor{rowcolor1}{RGB}{242, 246, 252}
\definecolor{rowcolor2}{RGB}{255, 255, 255}

\newtheorem{theorem}{Theorem}

\newtheorem{lemma}{Lemma}

\newtheorem{proposition}{Proposition}

\newtheorem{corollary}{Corollary}

\newtheorem{property}{Property}

\newtheorem{remark}{Remark}

\newtheorem{claim}{Claim}

\newcolumntype{A}{>{\hsize=0.95\hsize\raggedright\arraybackslash}X}  
\newcolumntype{B}{>{\hsize=0.6\hsize\centering\arraybackslash}X}   
\newcolumntype{C}{>{\hsize=1.7\hsize\centering\arraybackslash}X}   
\newcolumntype{D}{>{\hsize=0.75\hsize\centering\arraybackslash}X}  

 \usepackage[ bookmarks=false]{hyperref}
\hypersetup{
  colorlinks=false,
  urlcolor=black,
  linkcolor=black,
  citecolor=black,
  pdfborder={0 0 0}
}
\def\BibTeX{{\rm B\kern-.05em{\sc i\kern-.025em b}\kern-.08em
    T\kern-.1667em\lower.7ex\hbox{E}\kern-.125emX}}

\begin{document}
 
 \pagestyle{empty}

\title{\huge Dual-Attention and Adversarial Transfer Networks for Sim-to-Real Cross-Orientation Wireless Sensing}

\author
{\IEEEauthorblockN{Linfeng Du\IEEEauthorrefmark{1}, Kehan Wu\IEEEauthorrefmark{1}, Tong Zhang and Rui Wang}\vspace{-0.20cm}
\thanks{\IEEEauthorrefmark{1}Both authors contributed equally to this work.}
\thanks{L. Du is with National Graduate College for Engineers, Southern University of Science and Technology, Shenzhen, 518055, China (e-mail: dulf2024@mail.sustech.edu.cn).}
\thanks{K. Wu is with National Graduate College for Engineers, Southern University of Science and Technology, Shenzhen, 518055, China (e-mail: wukh2024@mail.sustech.edu.cn).}
\thanks{T. Zhang is with Guangdong Provincial Key Laboratory of Aerospace Communication and Networking Technology, Harbin Institute of Technology, Shenzhen, 518071, China (e-mail: tongzhang@hit.edu.cn).}
\thanks{R. Wang is with the Southern University of Science and Technology, Shenzhen, 518055, China (e-mail: wangr@sustech.edu.cn).}

\thanks{The source code is available at: {\color{cyan} \url{https://github.com/DuLF131/S2Msense}}}

}
\maketitle

\begin{abstract}
Millimeter-wave human activity recognition suffers significant performance degradation when the user's orientation changes relative to the sensing system, yet collecting labeled multi-orientation data is labor-intensive and costly. To eliminate the need for exhaustive multi-orientation measured data, we develop a physics-guided simulator that synthesizes orientation-diverse wireless training data from single-orientation motion. Specifically, to suppress orientation-induced feature variations, we propose a dual-attention network that extracts activity-discriminative and orientation-robust representations from dual-link Doppler spectrograms. To bridge the simulation-to-reality gap, we introduce an adversarial unsupervised transfer learning mechanism that aligns feature distributions using only a small number of unlabeled target-domain samples. The S2M-Sense platform shows high fidelity in reproducing real-world signatures, validated against $60.48$~GHz mmWave measured data with an average structural similarity index measure (SSIM) of $0.84$ between simulated and measured Doppler spectrograms across all $4$ activities and $4$ orientations.  Experimental results show that S2M-Sense achieves $88.33$\% recognition accuracy using only the dual-link multi-orientation simulated dataset, which improves to $95$\% after simulation-to-reality transfer learning with as few as $16$ unlabeled measured samples. Both cases with and without transfer learning outperform state-of-the-art cross-domain sensing methods. 
\end{abstract}


\section{Introduction}
Recognition of human activity has attracted increasing attention for smart homes, healthcare monitoring, and human-computer interaction. Camera-based methods \cite{surek2023video, bukht2025review} and wearable systems \cite{contoli2024energy, khalifa2017harke} can achieve strong performance, but they raise privacy concerns or require users to carry dedicated devices. Recent studies leverage commodity Wi-Fi signals for human activity recognition to address these problems. Nevertheless,  wireless sensing performance drops when the user's orientation changes, because orientation shifts strongly distort the wireless signal and increase intra-class variance \cite{virmani2017position,zhang2018crosssense,zou2018robust,zhang2021widar3,cao2025real, gu2022wigrunt, zhang2023imgfi, 10233699, ren2024caster}. 

Existing cross-domain wireless sensing methods fall into two categories. The first category synthesizes target-domain samples. For example, WiAG used a theoretically grounded translation function to generate virtual samples for any user position or orientation from a single configuration, eliminating the need for labeled target-domain data \cite{virmani2017position}. CrossSense used an offline-trained roaming model to generate synthetic training samples from one set of measurements, enabling cross-site Wi-Fi translation and large-scale sensing with less labeled target data \cite{zhang2018crosssense}. The second category learns features that are invariant across domains. For example, WiADG uses unsupervised adversarial domain adaptation to enable gesture recognition in new environments without labeled target data or retraining \cite{zou2018robust}, while Widar3.0 extracts domain-independent body-coordinate velocity profiles from Wi-Fi CSI for zero-effort cross-domain gesture recognition without new-domain data collection or retraining \cite{zhang2021widar3}. WiDual \cite{cao2025real} introduced a dual-task framework with collaborative learning, while WiGRUNT \cite{gu2022wigrunt} employed spatial--temporal dual attention to learn domain-invariant features. And ImgFi \cite{zhang2023imgfi} converted channel state information into images for CNN-based light recognition. However, all of these methods remain heavily dependent on measured multi-orientation data, and collecting labeled data for each domain is labor-intensive for real-world deployment, making cross-orientation recognition challenging.

To alleviate dependence on measured data, the simulation-to-reality approach has been proposed \cite{10233699,ren2024caster}, which simulate and recognize actions only under a single orientation. \cite{10233699} proposed a deep spectrogram network and a primitive-based autoregressive hybrid channel model for virtual human motion dataset generation, while \cite{ren2024caster} developed CASTER, which combines monocular-video motion capture with primitive-based ray tracing to simulate gesture-induced wireless channels. However, neither study addresses cross-orientation recognition.

 In this paper, we therefore propose S2M-Sense, a system-oriented platform for cross-orientation sensing that combines simulation-driven data generation, orientation-robust feature learning, and simulation-to-reality adaptation. Rather than collecting measurements under every orientation, S2M-Sense generates orientation-diverse training data from single-orientation depth-camera capture. We employ a passive sensing system using OFDM signals with the cross-ambiguity function to obtain measured data, that are used to validate the proposed platform. Our contributions are threefold. 1) We develop a high-fidelity simulation-driven data generation platform that expands single-orientation depth-camera capture into multi-orientation wireless training data. 2) We design a dual-attention network for dual-link Doppler spectrograms to better extract orientation-robust features. This design improves action feature extraction and reduces the drawbacks of traditional serial attention modules. 3) We combine simulation pretraining with adversarial unsupervised transfer learning and validate the $95$\% accuracy on measured $60$ GHz sensing data.

\section{Proposed Simulation-to-Reality Cross-Orientation Wireless Sensing Platform} 
\label{sec:system}
The proposed S2M-Sense enables unified human activity recognition across different orientations in real-world scenarios through passive sensing using OFDM signals. Specifically, it operates in a sensing setup with two near-orthogonal millimeter-wave communication links. When a target person is within the joint coverage of the two links and performs activities at different orientations, the system applies the cross-ambiguity function to the received surveillance and reference signals to obtain two Doppler spectrograms. These dual-link spectrograms are then fed into the proposed recognition net-work for cross-orientation activity classification. The network is first trained on the cross-orientation simulated dataset and adapted with a small amount of unlabeled measured data.

\begin{figure}[!t]
  \centering
  \includegraphics[height=0.25\textwidth]{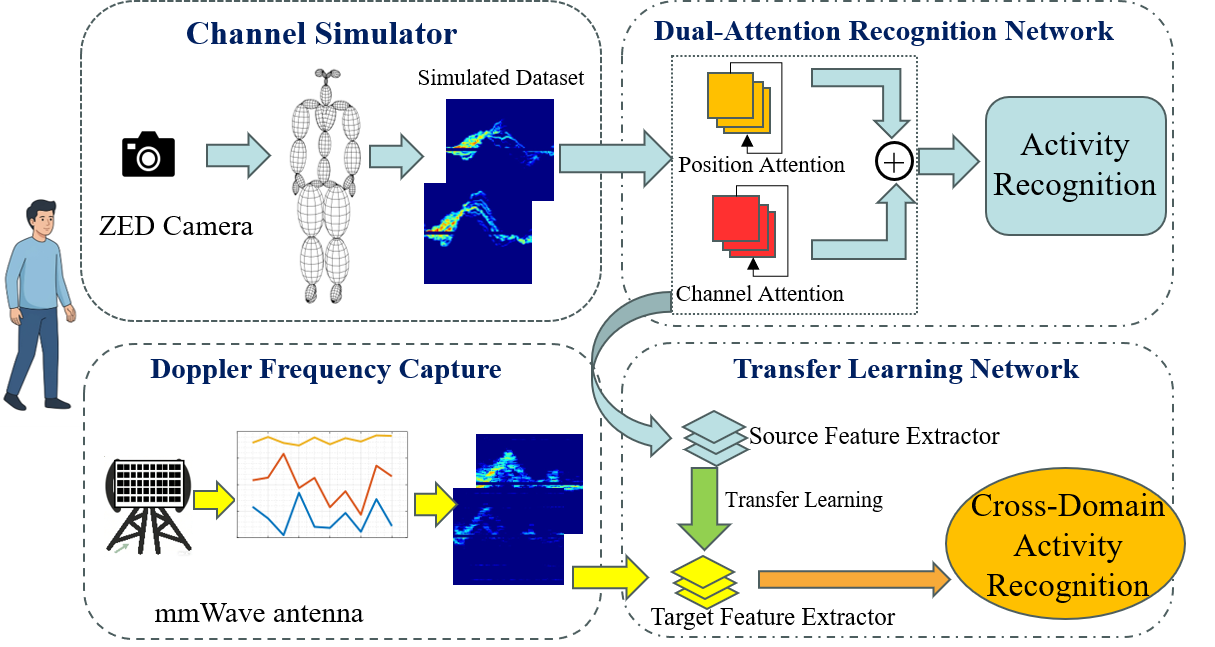}  
  \caption{Architecture of S2M-Sense.}
  \label{fig:overview}
  \vspace{-0.1cm}
\end{figure}

As shown in Fig. \ref{fig:overview}, S2M-Sense consists of four components: channel-simulator-driven data generation, measured Doppler capture, dual-attention feature extraction, and sim-to-real transfer learning. The channel simulator generates cross-orientation simulated data from depth-camera videos, while measured OFDM signals from two orthogonal links are converted into Doppler spectrograms. The dual-attention network then extracts robust activity features from the simulated dual-link spectrograms. Finally, adversarial unsupervised transfer learning uses a small amount of unlabeled measured data to align the simulated and measured domains, thereby improving recognition in real-world scenarios.

The channel simulator maps human motion to wireless measurements. The complete activity is decomposed into a sequence of depth-camera frames. In each frame, 34 body keypoints are estimated by a depth camera, and adjacent keypoints are connected by ellipsoidal primitives to approximate the human body. Under the quasi-static assumption, the CIR at time $t$ is written as
\begin{equation}
    h(t,\tau)=u(t,\tau)+v(\tau),
\end{equation}
where $u(t,\tau)$ is the target-related component and $v(\tau)$ captures LOS and static-scatterer paths.

To improve temporal continuity, the motion sequence is refined via cubic spline interpolation and smoothed with a One-Euro filter. This compensates for the limited frame rate and reduces sensitivity to abrupt keypoint variations, yielding smoother trajectories and more realistic synthetic spectrograms.

The channel simulator is described in two parts. First, we compute the channel impulse response (CIR) at time $t$ for a fixed orientation and extend it to multi-orientation scenarios. Second, we model the target-unrelated CIR. The overall CIR is obtained by combining the two components, thereby constructing a cross-orientation mmWave simulation dataset.
\subsection{Target-Related CIR and Orientation Expansion}
\begin{figure*}[!t]
\centering
\includegraphics[width=0.9\textwidth]{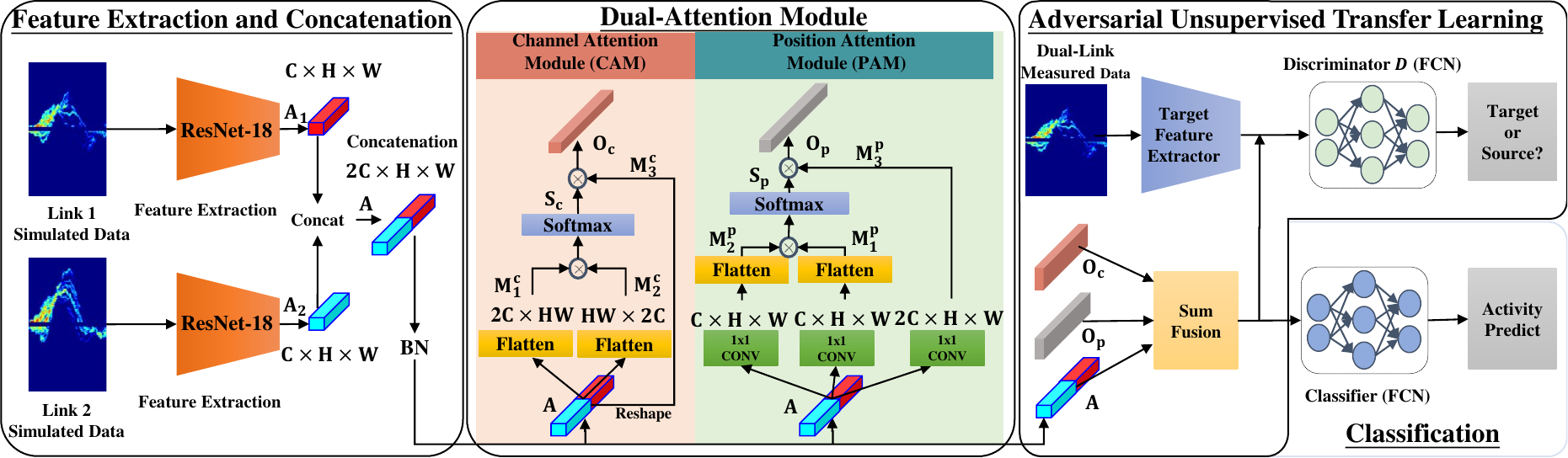}
\caption{Proposed Framework of the dual-attention network and adversarial unsupervised transfer learning.}
\label{fig: attention}
\end{figure*}
The contribution of the $k$-th ellipsoidal primitive to the bistatic channel impulse response (CIR) can be modeled as
\begin{equation}
u_k(t, \tau)
= \lambda
\sqrt{
\frac{
\sigma_k(t) G_k^{t}(t)G_k^{r}(t)
}{
(4\pi)^3 \left(R_k^{t}(t) R_k^{r}(t)\right)^2
}
}
e^{-j\phi_k(t)}
\delta\big(\tau - \tau_k(t)\big),
\end{equation}
where $G_k^{t}(t)$ and $G_k^{r}(t)$ denote the transmit and receive antenna gains for the $k$-th primitive, $R_k^{t}(t)$ and $R_k^{r}(t)$ are the distances from the transmitter and receiver to the primitive center, $\tau_k(t)=(R_k^{t}(t)+R_k^{r}(t))/c$ is the propagation delay, $\phi_k(t)=2\pi f_c \tau_k(t)$ is the phase shift, $\lambda$ is the wavelength, and $\delta(\cdot)$ is the Dirac delta function. The term $\sigma_k(t)$ denotes the bistatic radar cross section of the $k$-th primitive and is obtained following \cite{ren2024caster}.

Because the system uses a Sivers linear mmWave array antenna, the antenna gain is angle dependent. To model $G_k^t$ and $G_k^r$, we characterize this angular dependence using the array factor (AF) of a uniform linear array (ULA), i.e., $G^t_k=G^t*AF(\theta_k,\phi_k)$ and $G^r_k=G^r*AF(\theta_k,\phi_k)$, where $\theta_k$ and $\phi_k$ denote the azimuth and elevation angles of the reflector relative to the beam direction for the $k$-th primitive. The array factor of a ULA with $N$ elements is given by:

\begin{equation}
AF(\theta_k,\phi_k) = \sum_{n=0}^{N-1} e^{j n \frac{2\pi}{\lambda} d \sin\theta_k \cos\phi_k},
\label{AF}
\end{equation}
where $d$ is the inter-element spacing. This formulation follows classical antenna array theory  \cite{Balanis2005}.

To generate multi-orientation simulated data, we rotate the original keypoints around a vertical axis passing through the waist using a transformation matrix $\mathbf{T}(\alpha)$. $\alpha$ represents the clockwise rotation angle to the target orientation.

At the $t$-th snapshot, the coordinates of the $i$-th keypoint are $\mathbf{I}_i(t) = [x_i(t), y_i(t), z_i(t)]^T$. The rotation axis $l$ is defined as the line passing through the waist keypoint $\mathbf{I}_2(0) = [x_2(0), y_2(0), z_2(0)]^T$ at the initial snapshot and perpendicular to the $X$-$Z$ plane.

The $i$-th keypoint is rotated around axis $l$ by an angle $\alpha$, resulting in the transformed coordinate $\mathbf{I}_i^\alpha(t) = [x_i^\alpha(t), y_i^\alpha(t), z_i^\alpha(t)]^T$. Because the rotation is performed in the $X$-$Z$ plane, the $y$-coordinate remains unchanged. The transformation can be expressed as
\begin{equation}
\begin{bmatrix}
\mathbf{I}_i^\alpha(t) \\
1
\end{bmatrix}
=
\mathbf{T}(\alpha)
\begin{bmatrix}
\mathbf{I}_i(t) \\
1
\end{bmatrix},
\end{equation}
where
\begin{equation}
\mathbf{T}(\alpha) =
\begin{bmatrix}
\cos\alpha & 0 & -\sin\alpha & c_x(1-\cos\alpha)+c_z\sin\alpha \\
0 & 1 & 0 & 0 \\
\sin\alpha & 0 & \cos\alpha & c_z(1-\cos\alpha)-c_x\sin\alpha \\
0 & 0 & 0 & 1
\end{bmatrix},
\end{equation}
with $\mathbf{c} = [c_x, c_z]^T = [x_2(0), z_2(0)]^T$.

Accordingly, the set of transformed 3D keypoints at the $t$-th snapshot is given by
\begin{equation}
\mathbb{I}^\alpha(t) = \left\{ \mathbf{I}_i^\alpha(t) \mid i = 1, 2, \ldots, 34 \right\}.
\end{equation}

The corresponding channel impulse response is computed from the generated keypoint sequences at the target orientation using the ellipsoidal primitive-based ray-tracing model.
\subsection{Target-Unrelated CIR}

The target-unrelated component is composed of LOS and NLOS paths from static scatterers:
\begin{equation}
    v(\tau)=v_{\text{LOS}}(\tau)+v_{\text{NLOS}}(\tau).
\end{equation}
The LOS term is
\begin{equation}
v_{\text{LOS}}(\tau)=\frac{\lambda \sqrt{G_{\text{LOS}}^tG_{\text{LOS}}^r}}{4\pi R_{\text{LOS}}}
e^{-j\phi_{\text{LOS}}}\delta(\tau-\tau_{\text{LOS}}),
\end{equation}
and the NLOS term sums the reflections of $N$ static objects:
\begin{equation}
v_{\text{NLOS}}(\tau)=\sum_{n=1}^{N}\lambda\sqrt{\frac{\sigma_n G_n^tG_n^r}{(4\pi)^3(R_n^tR_n^r)^2}}
e^{-j\phi_n}\delta(\tau-\tau_n).
\end{equation}
The antenna gain is modeled using (\ref{AF}). Doppler spectrograms are then obtained from the CIRs through Fourier analysis.

\section{Proposed Dual-Attention and Transfer Learning Networks}
\label{sec:network}
After obtaining the simulated channel impulse response, we apply Fourier transform to generate the simulated Doppler spectrum and construct the simulated dataset. Nevertheless, cross-orientation activity recognition remains challenging due to orientation-dependent variations and the sim-to-real gap.
\subsection{Dual-Attention Network}
To improve the recognition accuracy, as shown in Fig. \ref{fig: attention}, inspired by DANet \cite{fu2019dual}, we design a dual-attention module to extract activity-related features while reducing the impact of orientation changes. Specifically, position attention captures spatial dependencies to highlight action regions and suppress noise, whereas channel attention models inter-channel dependencies to emphasize informative channels and suppress orientation-sensitive channels. 

To capture richer motion information across orientations, we use the channel simulator to generate spectrograms from two orthogonal links and adopt ResNet-18 to extract robust representations $\mathbf{A_1}$ and $\mathbf{A_2}$, where $\mathbf{A_1}, \mathbf{A_2} \in \mathbb{R}^{C \times H \times W}$. We concatenate them along the channel dimension to obtain $\mathbf{A} = \operatorname{Concat}_c(\mathbf{A_1}, \mathbf{A_2})$, where $\mathbf{A} \in \mathbb{R}^{2C \times H \times W}$, and apply a batch normalization (BN) layer to normalize the feature distribution, stabilize optimization, and improve cross-link feature alignment. The normalized features are then fed into the position and channel attention modules to learn spatial correlations and channel-wise dependencies, enabling the model to focus on action-related features.

\emph{Position Attention Module}: For the position attention module, the feature $\mathbf{A}$ first passes through three convolutional layers to generate three new feature maps $\mathbf{M_1^p}$, $\mathbf{M_2^p}$, and $\mathbf{M_3^p}$, which are reshaped to $\mathbf{M_1^p}, \mathbf{M_2^p} \in \mathbb{R}^{C \times N}$ and $\mathbf{M_3^p} \in \mathbb{R}^{2C \times N}$, where $N = H \times W$. We then perform matrix multiplication between $\mathbf{M_1^p}$ and the transpose of $\mathbf{M_2^p}$, followed by a softmax layer, to obtain the position attention map $\mathbf{S_p} \in \mathbb{R}^{N \times N}$:
\begin{equation}
\mathbf{S_p} = \mathrm{softmax}\left(\mathbf{(M_2^p)^\top}\mathbf{M_1^p}\right).
\end{equation}
The position attention map captures long-range spatial dependencies by modeling correlations among all spatial positions, thereby highlighting motion-relevant regions and suppressing irrelevant backgrounds. Finally, we multiply $\mathbf{M_3^p}$ by $\mathbf{S_p^\top}$ and reshape the result into $\mathbb{R}^{2C \times H \times W}$ to obtain the output $\mathbf{O_p}$ of the position attention module:
\begin{equation}
\mathbf{O_p} = \mathbf{M_3^p}\mathbf{S_p^\top}.
\end{equation}
Here, $\mathbf{O_p}$ is the weighted feature of the position attention map $\mathbf{S_p}$ and the features $\mathbf{M_3^p}$, allowing selective aggregation of global context and enhancing the model's focus on activity-relevant information. Finally, a BN layer is applied to ensure stable training.

\emph{Channel Attention Module}: The channel attention module follows a similar procedure. The feature map $\mathbf{A}$ is flattened along two branches to obtain two feature maps $\mathbf{M_1^c}$ and $\mathbf{M_2^c}$, where $\mathbf{M_1^c}, \mathbf{M_2^c} \in \mathbb{R}^{2C \times N}$ and $N = H \times W$. These feature maps are multiplied and then passed through a softmax activation function to generate the channel attention map $\mathbf{S_c} \in \mathbb{R}^{2C \times 2C}$: 
\begin{equation}
\mathbf{S_c} = \mathrm{softmax}\left(\mathbf{M_1^c}\mathbf{(M_2^c)^\top}\right).
\end{equation}
The channel attention map models inter-channel dependencies, assigning higher weights to channels that are strongly correlated with the activity while suppressing irrelevant or background-related channels. The original feature map is then reshaped to $\mathbb{R}^{2C \times N}$ to obtain $\mathbf{M_3^c}$. We multiply the transpose of the channel attention map $\mathbf{S_c^\top}$ by $\mathbf{M_3^c}$, resulting in the output $\mathbf{O_c}$ of the channel attention module:
\begin{equation}
\mathbf{O_c} = \mathbf{S_c^\top}\mathbf{M_3^c}.
\end{equation}
$\mathbf{O_c}$ is the weighted feature of the channel attention map $\mathbf{S_c}$ and the features $\mathbf{M_3^c}$, allowing the model to capture more discriminative feature representations by exploring correlations across channels.

Finally, we introduce a residual connection by fusing $\mathbf{O_p}$, $\mathbf{O_c}$, and the original feature map $\mathbf{A}$ to obtain the final feature representation $\mathbf{O}$:
\begin{equation}
\mathbf{O} = a\cdot\mathbf{O_p} + b\cdot\mathbf{O_c} + \mathbf{A},
\end{equation}
where $a$ and $b$ are learnable parameters initialized to 0. We then feed $\mathbf{O}$ into a classifier composed of fully connected layers to predict the action label. The loss function is defined as
\begin{equation}
\mathcal{L} = -\frac{1}{N}\sum_{i=1}^{N}
\log \left(
\frac{e^{z_{i,y_i}}}{\sum_{j=1}^{C} e^{z_{i,j}}}
\right),
\end{equation}
where $N$ is the batch size, $C$ is the number of classes, $z_{i,j}$ denotes the logit for the $j$-th class of the $i$-th sample, $y_i$ is the ground-truth label, and $z_{i,y_i}$ denotes the logit corresponding to the ground-truth class of the $i$-th sample.

\subsection{Adversarial Unsupervised Transfer Learning Network}
To bridge the gap between simulated (source) and measured (target) data, we employ an adversarial approach inspired by ADDA \cite{tzeng2017adversarial}. As shown in Fig. \ref{fig: attention}, the source feature extractor ${f_{{s}}}(\cdot)$ and classifier are first pre-trained on simulated data. Let ${D}(\cdot)$ denote the domain discriminator, which consists of a  $7\times7$ convolutional layer and three fully connected layers. The labeled simulated dataset forms the source domain, where ${\mathcal{X}_{s}}$ and ${\mathcal{Y}_{s}}$ denote the simulated samples and their corresponding labels. A small set of unlabeled measured samples forms the target domain, denoted by ${\mathcal{X}_{t}}$. The target feature extractor ${f_{t}}(\cdot)$ is initialized with the same parameters as ${f_{s}}(\cdot)$. Training then alternates between the following two steps:

First, to optimize the discriminator $D$, we fix the feature extractors and train $D$ to distinguish source-domain features $f_s(x_s)$ from target-domain features $f_t(x_t)$ by minimizing $\mathcal{L}_{D}$. In this stage, the discriminator is encouraged to identify whether an input feature originates from the source domain or the target domain. The corresponding loss is defined as
\begin{align}
    \nonumber
    \min_{D} \mathcal{L}_{D}(\mathcal{X}_s, \mathcal{X}_t, f_s, f_t) = - \mathbb{E}_{x_s \in \mathcal{X}_s} \left[ \log D(f_s(x_s)) \right]\\
-\mathbb{E}_{x_t \in \mathcal{X}_t} \left[ \log \left(1 - D(f_t(x_t)) \right) \right]. 
\end{align}

Second, to optimize the target feature extractor, we fix $D$ and fine-tune $f_t$ to confuse the discriminator by minimizing $\mathcal{L}_{f}$. As training proceeds, $f_t$ is encouraged to generate target-domain features that are increasingly indistinguishable from source-domain features, thereby reducing the domain discrepancy. The corresponding loss is defined as
\begin{equation} 
\min_{f_t} \mathcal{L}_{f}(\mathcal{X}_s, \mathcal{X}_t, D) = - \mathbb{E}_{x_t \in \mathcal{X}_t} \left[ \log D(f_t(x_t)) \right] .
\end{equation}

Through alternating optimization, the output features $f_t(x_t)$ gradually align with the source-domain feature distribution, enabling the transferred model to recognize actions effectively from measured data.

\section{Experiments}
\label{sec:experiment}

\begin{figure*}[t]
    \centering
    \begin{minipage}[t]{0.24\textwidth}
        \centering
        \includegraphics[width=0.8\linewidth]{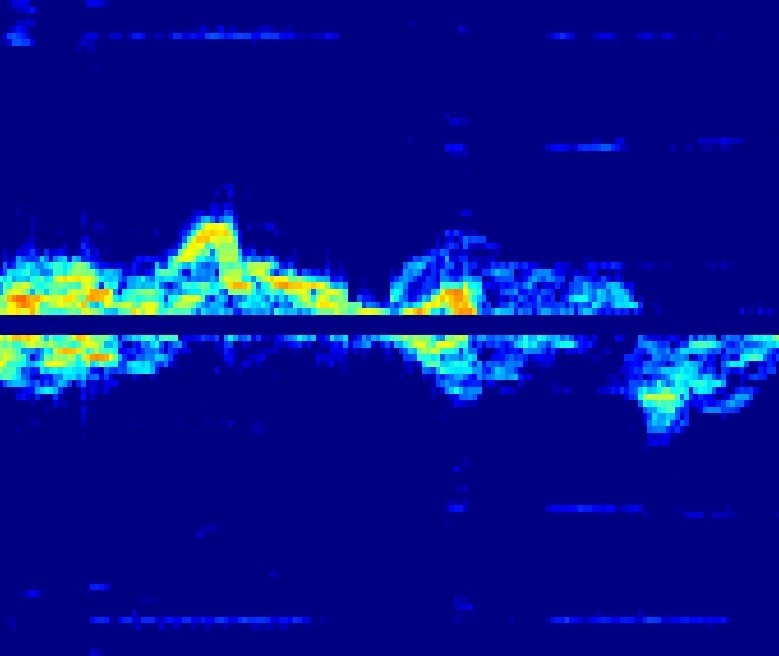}\\[-0.25em]
        (a) Measured, sit
    \end{minipage}
    \hfill
    \begin{minipage}[t]{0.24\textwidth}
        \centering
        \includegraphics[width=0.8\linewidth]{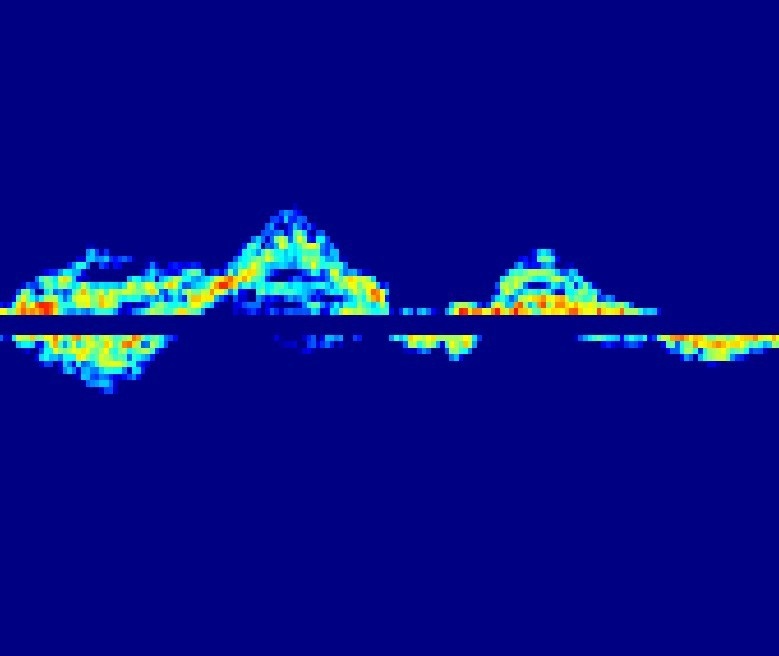}\\[-0.25em]
        (b) Simulated, sit
    \end{minipage}
    \hfill
    \begin{minipage}[t]{0.24\textwidth}
        \centering
        \includegraphics[width=0.8\linewidth]{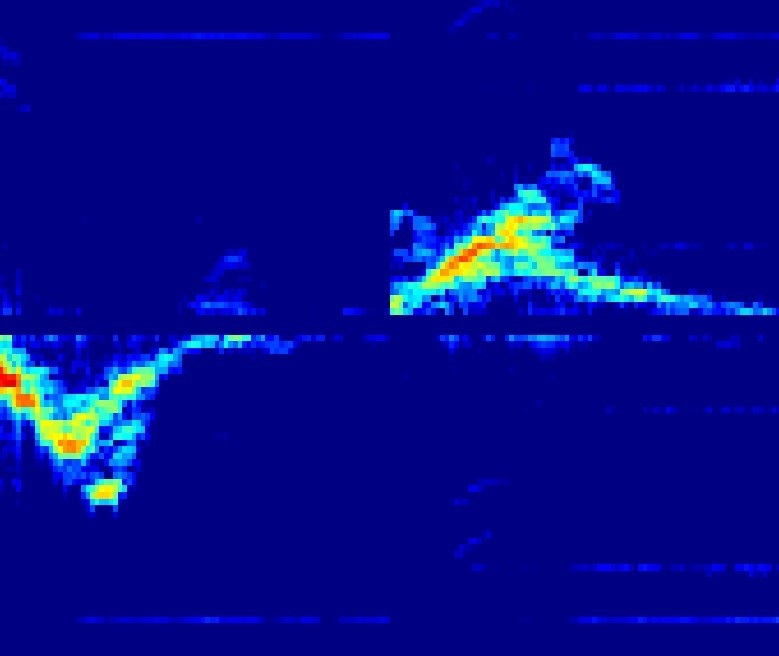}\\[-0.25em]
        (c) Measured, step
    \end{minipage}
    \hfill
    \begin{minipage}[t]{0.24\textwidth}
        \centering
        \includegraphics[width=0.8\linewidth]{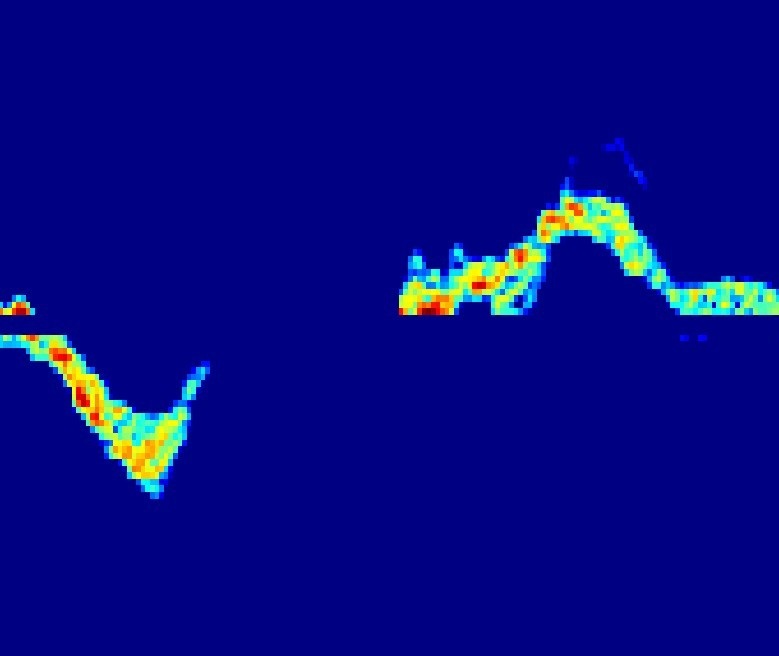}\\[-0.25em]
        (d) Simulated, step
    \end{minipage}

    \vspace{0.15em}

    \begin{minipage}[t]{0.24\textwidth}
        \centering
        \includegraphics[width=0.8\linewidth]{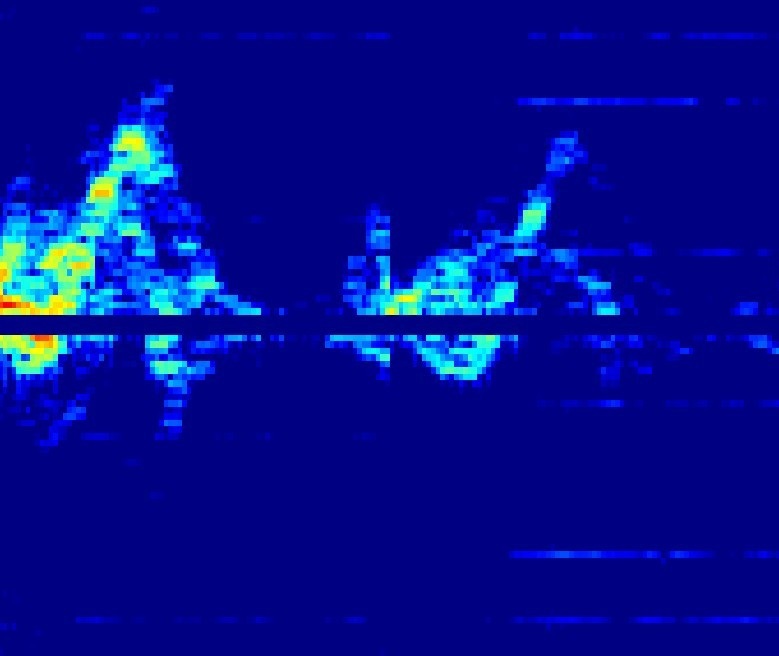}\\[-0.25em]
        (e) Measured, fall
    \end{minipage}
    \hfill
    \begin{minipage}[t]{0.24\textwidth}
        \centering
        \includegraphics[width=0.8\linewidth]{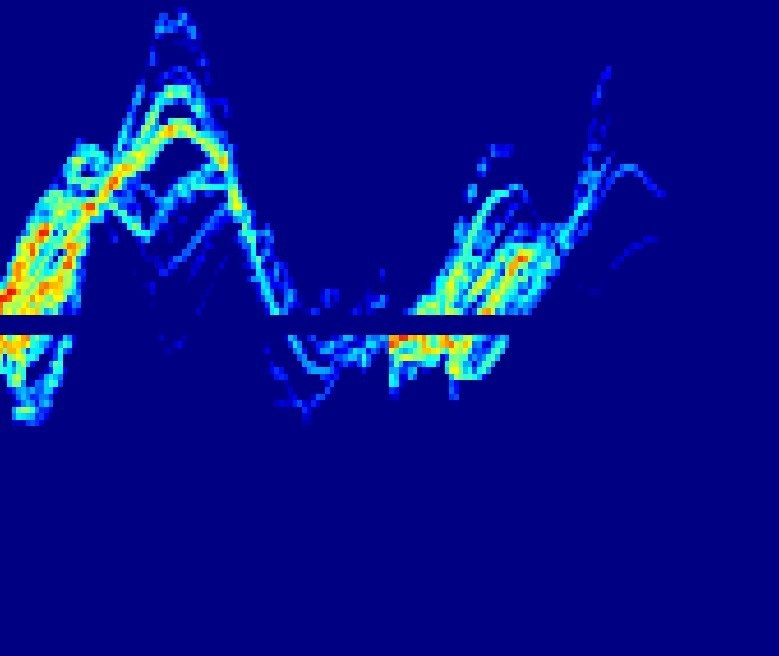}\\[-0.25em]
        (f) Simulated, fall
    \end{minipage}
    \hfill
    \begin{minipage}[t]{0.24\textwidth}
        \centering
        \includegraphics[width=0.8\linewidth]{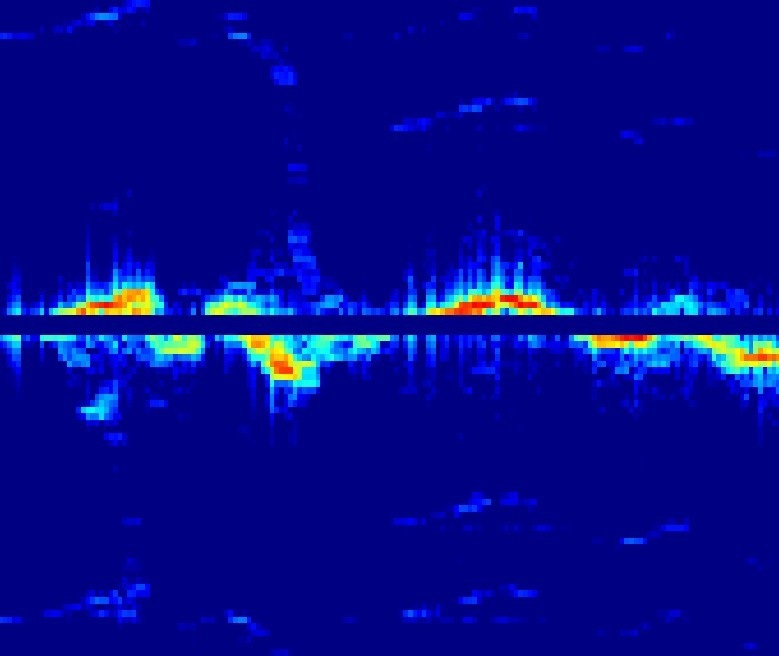}\\[-0.25em]
        (g) Measured, kick
    \end{minipage}
    \hfill
    \begin{minipage}[t]{0.24\textwidth}
        \centering
        \includegraphics[width=0.8\linewidth]{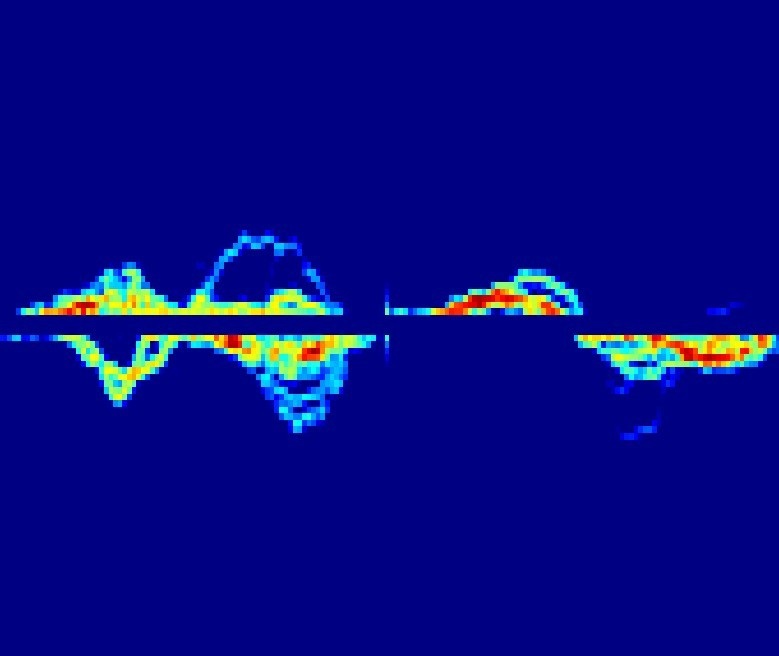}\\[-0.25em]
        (h) Simulated, kick
    \end{minipage}

   \caption{Measured vs. simulated Doppler spectrograms for four actions (left: measured, right: simulated).}
    \label{fig:real_sim}
\end{figure*}
\subsection{Settings and Baselines}
The sensing platform operates at 60.48~GHz. An NI USRP-2954R generates the IF signal, which is upconverted and transmitted by a Sivers phased-array antenna. At the receiver, two synchronized phased arrays capture the reference and surveillance channels. The waveform contains a training sequence and an OFDM payload, while the reference array synchronizes the surveillance array to keep the two channels phase-aligned for acquisition and Doppler processing. As shown in Fig.~\ref{fig:layout}, the two links are near-orthogonal and provide complementary views of motion from different orientations. The ZED coordinate system is used as the reference origin. For the first link, the transmitter (Tx) and receiver (Rx) are located at $[-0.4\,\mathrm{m}, -0.1\,\mathrm{m}, 0\,\mathrm{m}]$ and $[-0.4\,\mathrm{m}, 0.65\,\mathrm{m}, 0\,\mathrm{m}]$, respectively. For the second link, the Tx and Rx are positioned at $[2.1\,\mathrm{m}, 2.55\,\mathrm{m}, 0\,\mathrm{m}]$ and $[1\,\mathrm{m}, 2.55\,\mathrm{m}, 0\,\mathrm{m}]$, respectively. The transmitter is placed 2\,m from the target, while static scatterers are randomly sampled in the surrounding area to model target-unrelated paths. Using this passive sensing system and the cross-ambiguity function (CAF), we obtain 480 measured spectrograms.

The dataset includes four activities: \emph{step}, \emph{fall}, \emph{sit}, and \emph{kick}. For simulation, each action is recorded with a ZED 2i depth camera at 30 fps for 2\,s and then interpolated to 2000 fps. From 10 single-orientation videos collected from three volunteers, we generate 3000 spectrograms by randomly sampling orientations over $0^\circ$--$360^\circ$, covering the full azimuth range. For evaluation, measured data are collected from three different volunteers at four orientations, yielding 480 spectrograms. The test set covers viewpoints of $45^\circ$, $135^\circ$, $225^\circ$, and $315^\circ$. Specifically, the simulated training set uses volunteers A, B, and C, while the measured test set uses volunteers D, E, and F. This disjoint subject split reduces identity leakage and better reflects practical deployment. We train the model using Adam with a learning rate of $10^{-4}$ for 200 epochs and a batch size of 16. We compare the proposed method with three representative cross-orientation sensing baselines, all evaluated using the same training and testing datasets to ensure a fair comparison:
\begin{figure}[!t]
  \centering
  \includegraphics[width=0.36\textwidth]{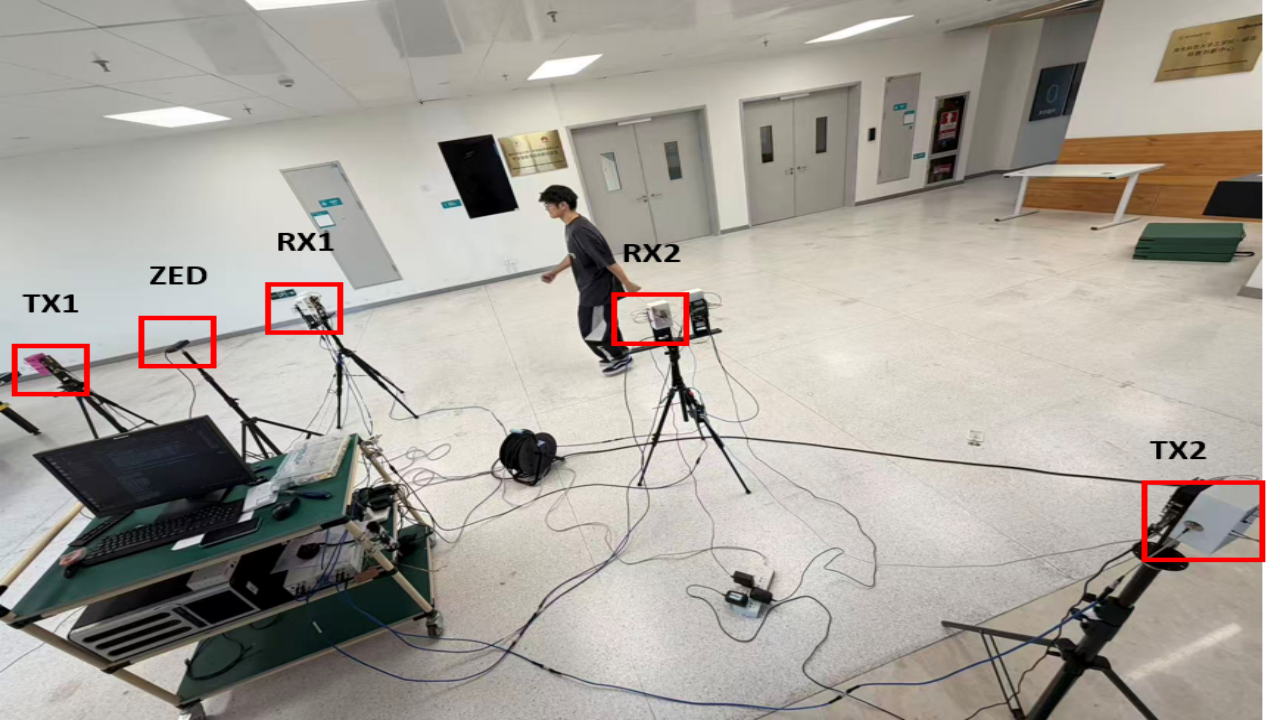}
  
  \caption{Experimental layout of the 60\,GHz sensing system.}
  \label{fig:layout}
\end{figure}

\textit{\textbf{WiDual} \cite{cao2025real}:} WiDual is a cross-domain dual-task framework for gesture recognition and user identification based on Wi-Fi CSI. It converts CSI into image-like representations and uses attention-based networks to learn discriminative features.

\textit{\textbf{WiGRUNT} \cite{gu2022wigrunt}:} WiGRUNT is a gesture recognition framework with a spatial-temporal dual-attention mechanism. Built on a residual backbone, it emphasizes domain-invariant features in both the spatial and temporal dimensions of CSI data to improve cross-domain generalization.

\textit{\textbf{ImgFi} \cite{zhang2023imgfi}:} ImgFi is a lightweight CSI-based activity recognition framework. It transforms CSI into image representations and applies a compact CNN for classification, balancing recognition accuracy and computational cost.

\subsection{Experiment Studies}
Fig. \ref{fig:real_sim} shows that the simulator preserves the dominant Doppler structures of each activity across orientations, although the measured spectrograms still contain clutter and amplitude distortions. As shown in Table \ref{tab:ssim_results}, we use the Structural Similarity Index Measure (SSIM) to quantify the similarity between simulated and measured spectrograms by comparing samples of the same activity class across the two datasets. SSIM ranges from 0 to 1, where 1 indicates identical images, and 0 indicates no similarity. The average simulated spectrograms achieve SSIM values above 0.84, indicating that the proposed pipeline reproduces the main measured patterns with high fidelity.
\begin{table}[t]
      \centering
      \caption{SSIM results under different orientations.}
      \label{tab:ssim_results}
      \small
      \renewcommand{\arraystretch}{1.05}
      \setlength{\tabcolsep}{6pt}
      
        \begin{tabular}{ccccc}
            \toprule
            Action & $45^\circ$ & $135^\circ$ & $225^\circ$ & $315^\circ$ \\
            \midrule
            Sit  & 0.8547 & 0.8884 & 0.8709 & 0.8812 \\
            Step & 0.8518 & 0.8693 & 0.8691 & 0.8796 \\
            Fall & 0.8020 & 0.8575 & 0.8332 & 0.7852 \\
            Kick & 0.8309 & 0.8460 & 0.8131 & 0.8088 \\
            \bottomrule
        \end{tabular}
    
  \end{table}
    
\begin{figure*}[t]
    \centering

    \begin{minipage}[t]{0.24\textwidth}
        \centering
        \includegraphics[width=0.9\linewidth]{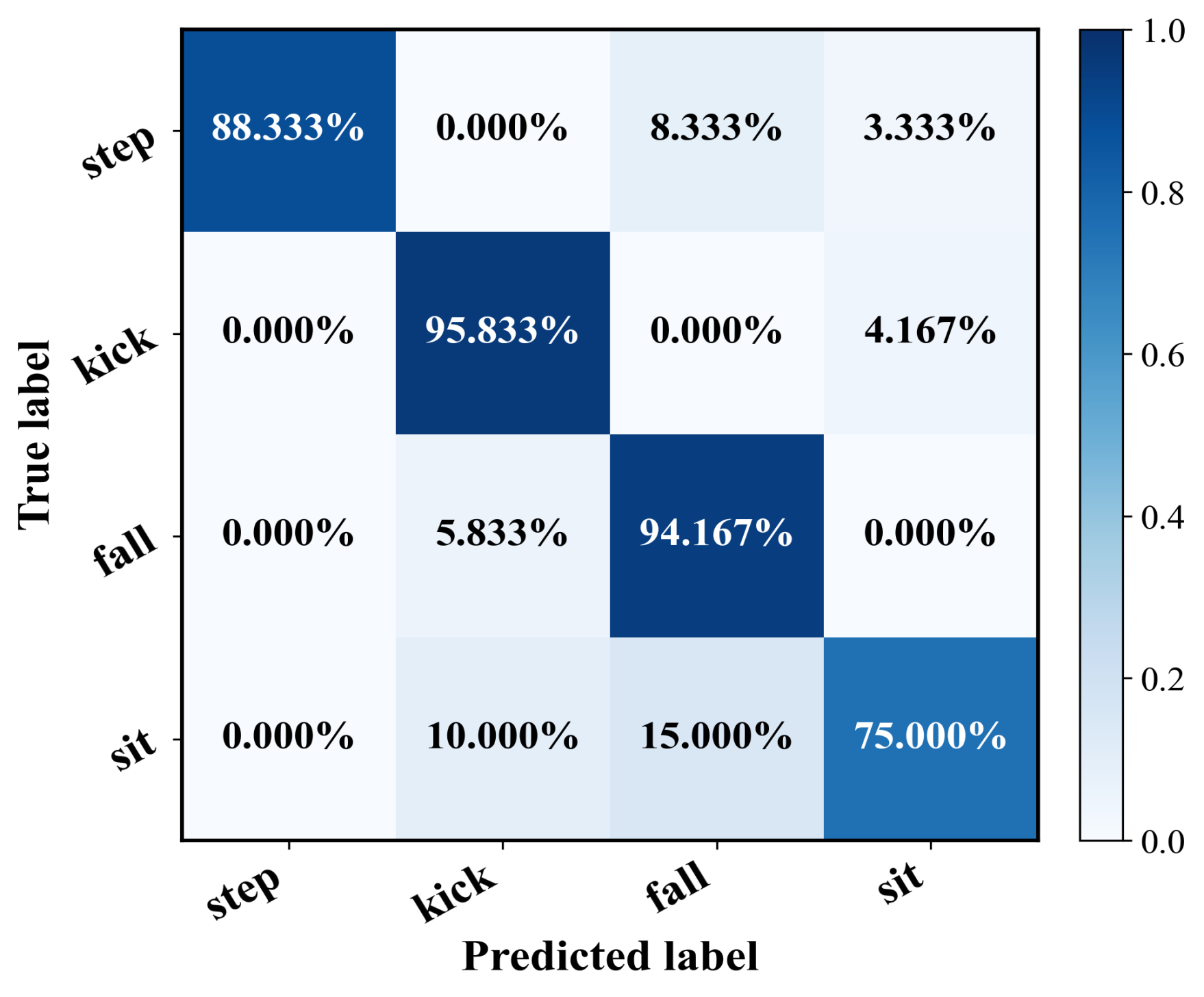}\\[-0.25em]
        (a) Confusion matrix, before transfer
    \end{minipage}
    \hfill
    \begin{minipage}[t]{0.24\textwidth}
        \centering
        \includegraphics[width=0.9\linewidth]{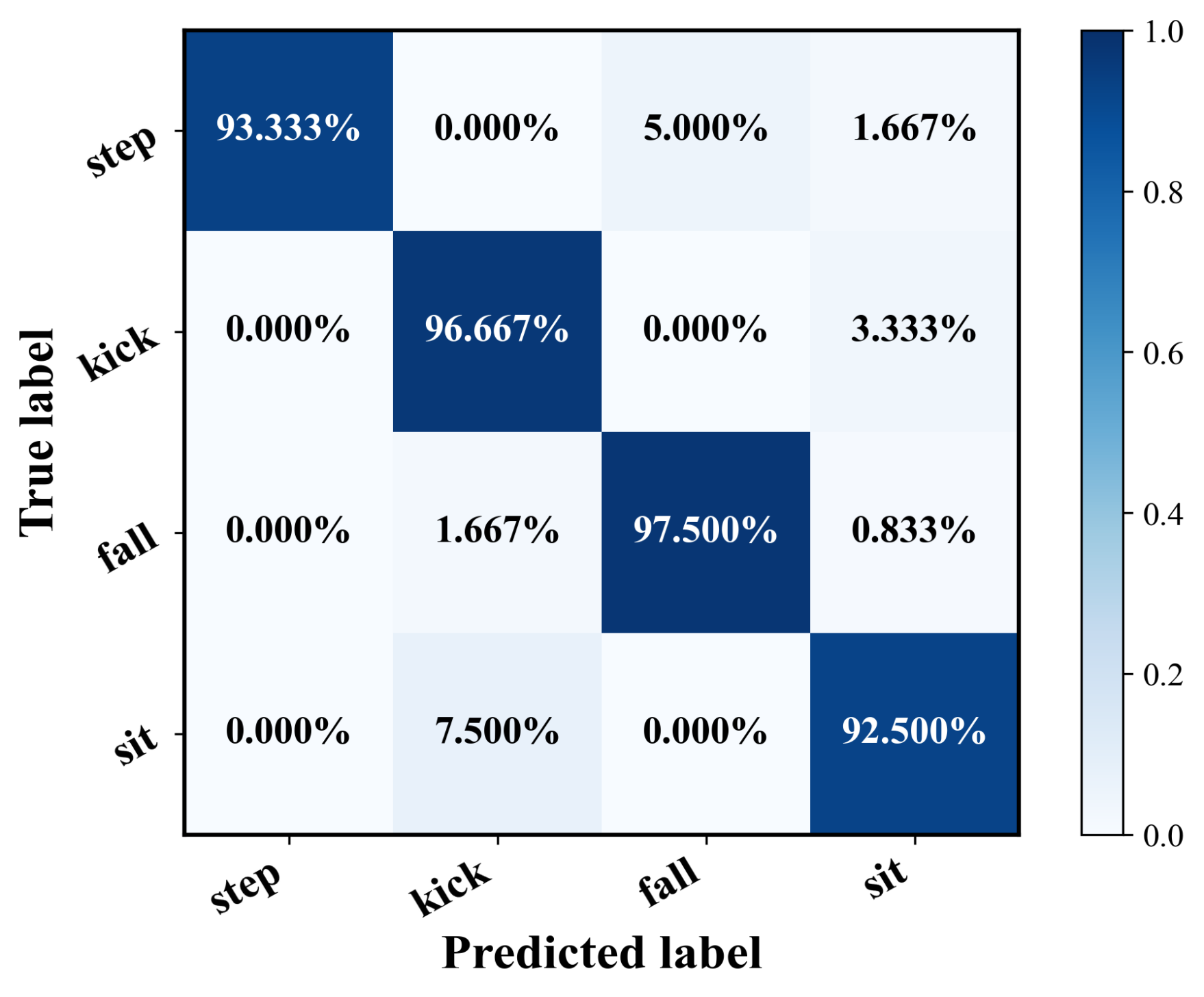}\\[-0.25em]
        (b) Confusion matrix, after transfer
    \end{minipage}
    \hfill
    \begin{minipage}[t]{0.24\textwidth}
        \centering
        \includegraphics[width=0.9\linewidth]{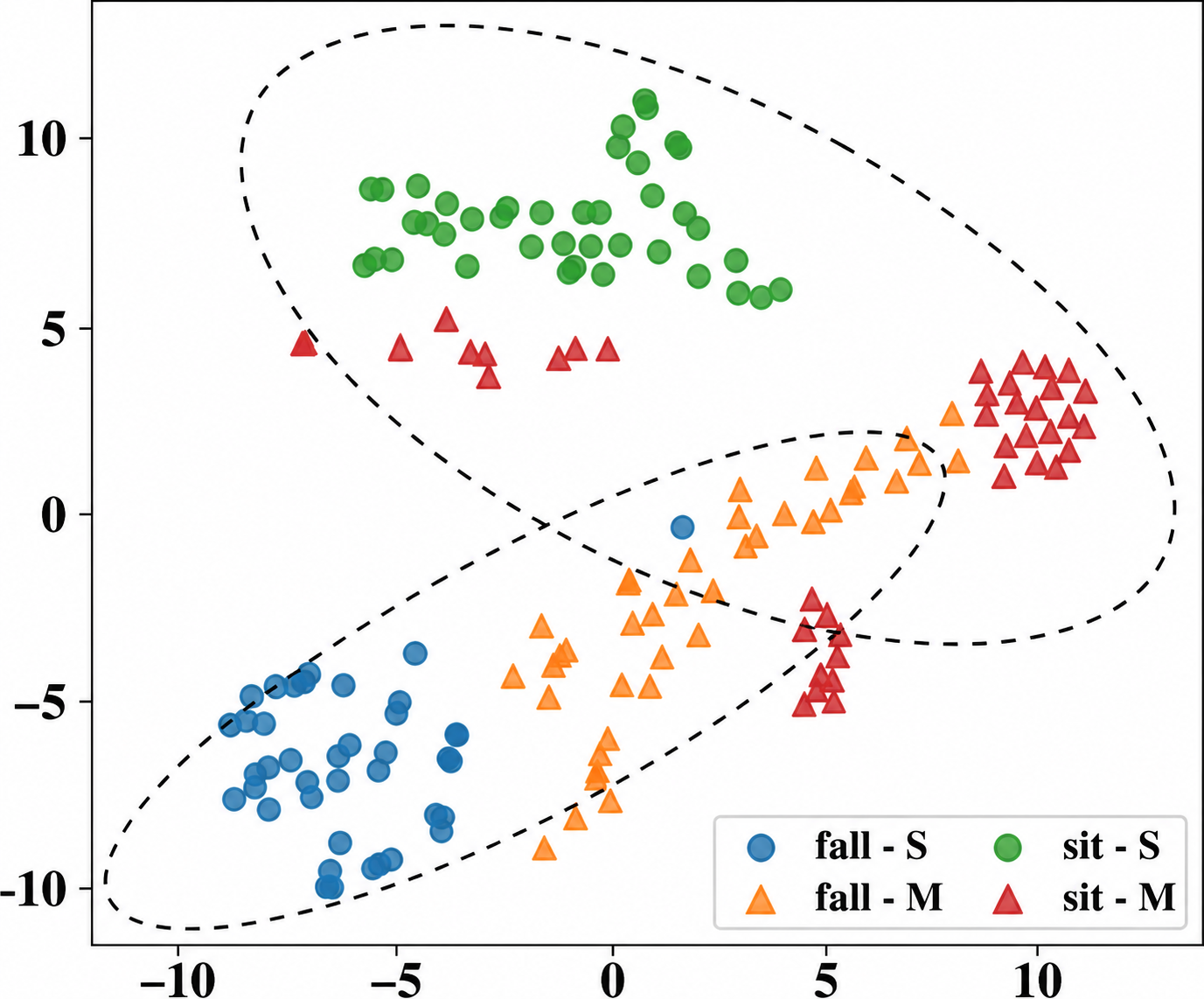}\\[-0.25em]
        (c) t-SNE, before transfer
    \end{minipage}
    \hfill
    \begin{minipage}[t]{0.24\textwidth}
        \centering
        \includegraphics[width=0.9\linewidth]{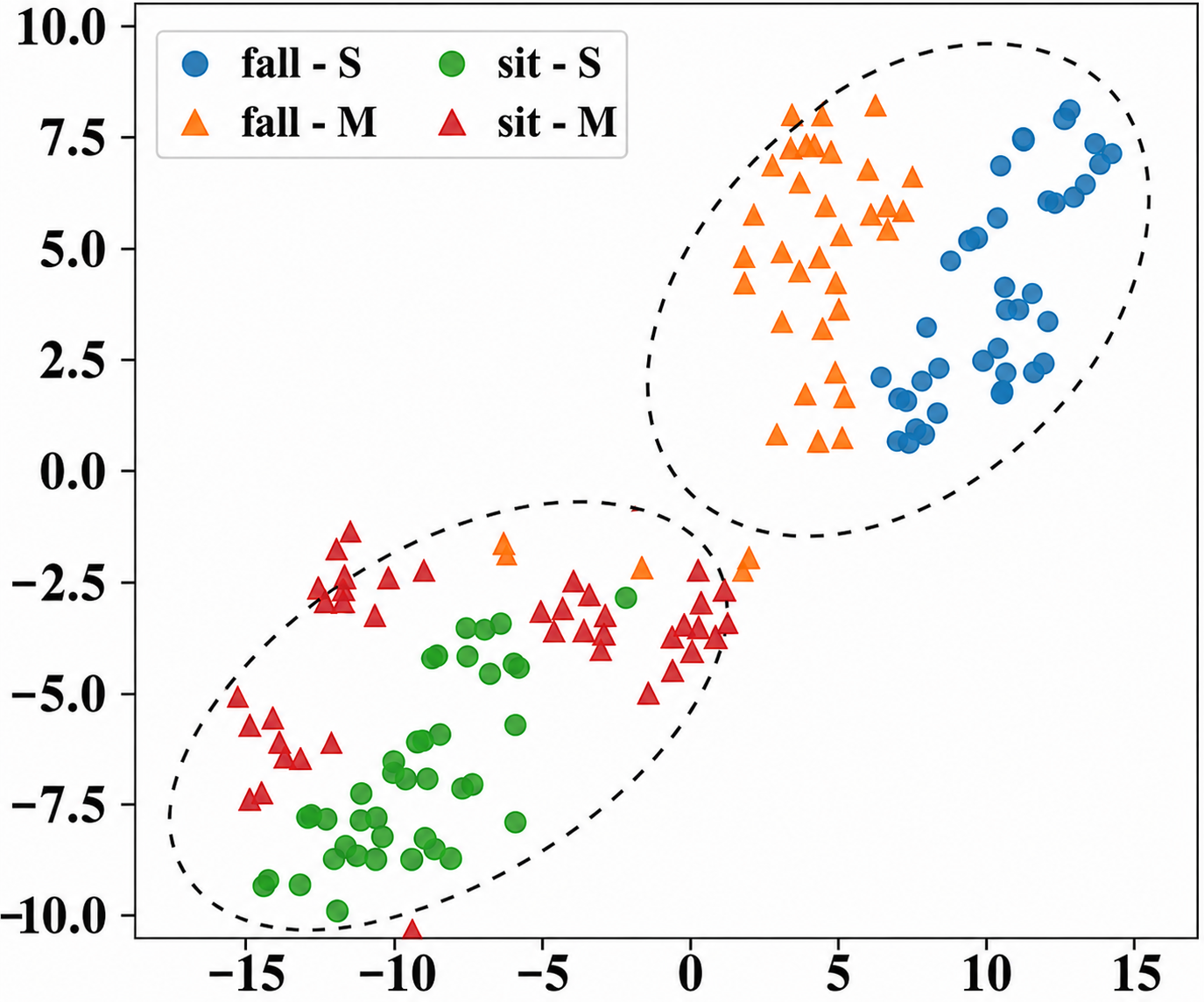}\\[-0.25em]
        (d) t-SNE, after transfer
    \end{minipage}    
    \caption{Confusion matrices and t-SNE visualizations before and after transfer.}
    \vspace{-0.4cm}
    \label{FG}
\end{figure*}
Fig. \ref{FG} adapts the source model using 16 unlabeled measured samples drawn from four directions. After adversarial transfer, the recognition accuracy rises from 88.33\% to 95\%, showing that even a small amount of unlabeled target data can effectively reduce the simulation-to-reality gap. In other words, the simulator provides a strong initialization, while adversarial adaptation further aligns the feature distributions with little deployment overhead. The t-SNE visualizations of \emph{fall} and \emph{sit} in Fig. \ref{FG} show better feature alignment after transfer, confirming that transfer learning effectively reduces the discrepancy between the simulated (S) and measured (M) domains.

\begin{figure}[t]
  \centering
  \includegraphics[width=0.46\textwidth]{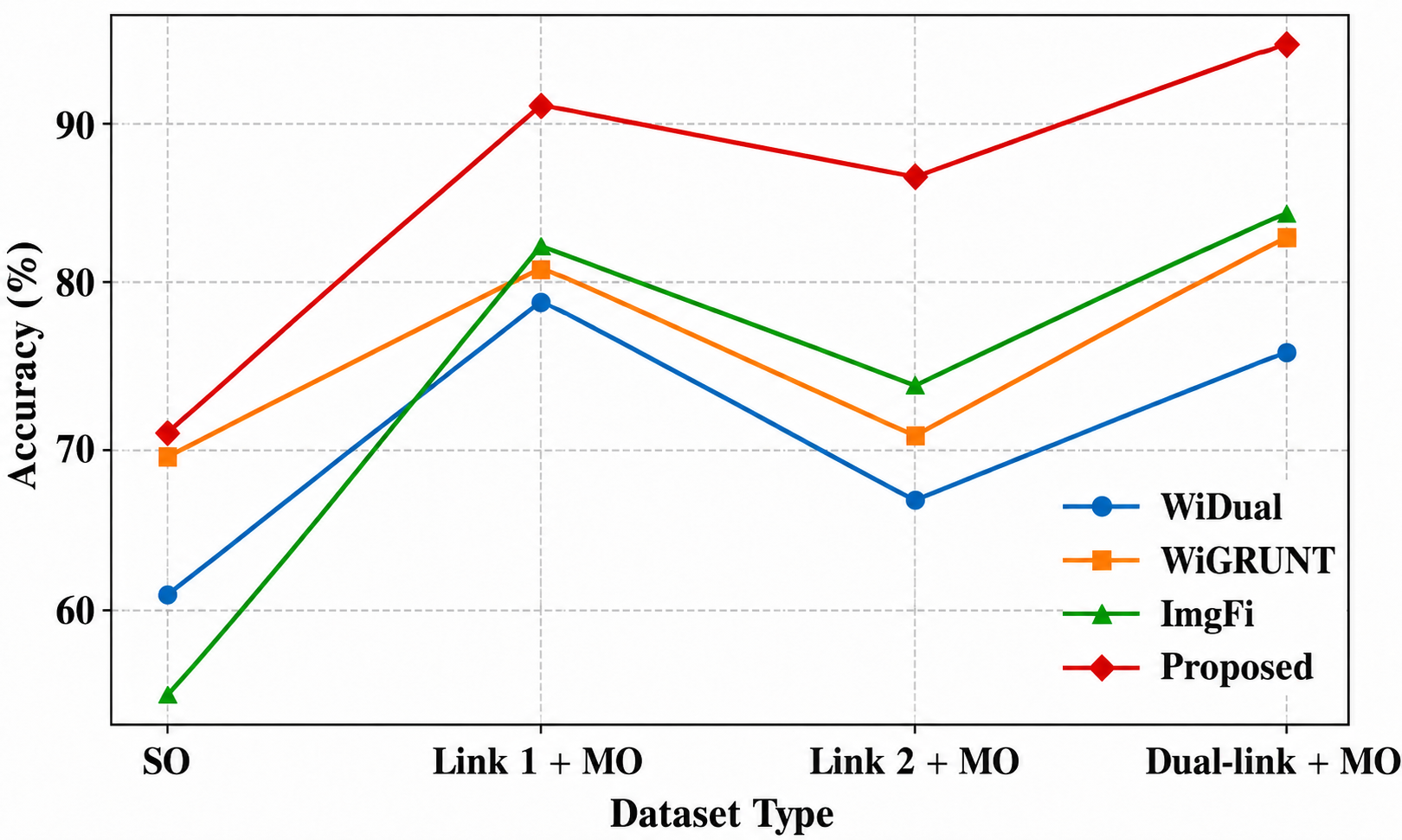}
  
  \caption{Accuracy comparison across various datasets.} 
  \label{fig:methods}
\end{figure}


Fig. \ref{fig:methods} compares the proposed method with state-of-the-art cross-domain recognition baselines. For a fair comparison, we retrain all baselines on four types of simulated datasets: single-orientation (SO), the first link with multi-orientation (Link 1 + MO), the second link with multi-orientation (Link 2 + MO), and dual-link with multi-orientation (Dual-link + MO), and then evaluate them on the measured test set. Our method consistently outperforms all baselines across the four datasets. The best performance is achieved on the dual-link multi-orientation dataset, reaching an accuracy of 95\%, which benefits from richer motion information. Using the best results of each baseline across the four datasets, our method attains 95\% accuracy, surpassing WiDual (78.96\%), WiGRUNT (81.04\%), and ImgFi (84.38\%). Unlike these baselines, the proposed system relies mainly on simulated data and requires only a small amount of unlabeled measured data for adaptation, greatly reducing data collection cost while maintaining high accuracy. In addition, the dual-attention module learns more robust cross-orientation representations by directly modeling feature responses, rather than using manually designed pooling operations. Its position and channel attention branches work in parallel with a residual connection, which improves feature and preserves attention effectiveness in deeper layers.

\section{Conclusion}
\label{sec:conclusion}
 This paper presented S2M-Sense, a simulation-driven platform for cross-orientation wireless activity recognition. Combining channel simulator, dual-attention networks, and adversarial transfer learning, it reduces reliance on measured data while achieving $88.33$\% accuracy with simulation-only training and $95$\% after adaptation with minimal unlabeled real data. Future work will target complex indoor and multi-person scenarios.

\bibliographystyle{ieeetr} 
\bibliography{SIMAR4}
\end{document}